\documentclass[a4paper,twocolumn,twoside,10pt]{ranlp}

\usepackage{graphicx}
\usepackage{mathrsfs}
\usepackage{amssymb,amsmath}
\usepackage{cyrillic}

\newcommand{\MUC}{\textsc{muc}}
\newcommand{\NER}{\textsc{ner}}
\newcommand{\ANNIE}{\textsc{annie}}
\newcommand{\CRF}{\textsc{crf}}

\newcommand{\MI}{\textsc{mi}}

\usepackage[usenames]{color}

\begin{document}

\title{\textbf{Feature-Rich Named Entity Recognition for Bulgarian
               Using Conditional Random Fields}}

\author{
Georgi Georgiev\footnote{Ontotext AD, 135 Tsarigradsko Ch., Sofia 1784, Bulgaria}\\
\email{georgi.georgiev}{ontotext.com}
\and
Preslav Nakov\footnote{Department of Computer Science, National University of Singapore, 13 Computing Drive, Singapore 117417} $^{\S}$\\
\email{nakov}{comp.nus.edu.sg}
\and
Kuzman Ganchev\footnote{Department of Computer and Information Science, University of Pennsylvania, Philadelphia, PA, USA}\\
\email{kuzman}{cis.upenn.ed}
\and
Petya Osenova\footnote{Linguistic Modelling Laboratory, Institute for Parallel Processing, Bulgarian Academy of Sciences,
25A Acad. G. Bonchev St., 1113 Sofia, Bulgaria}\\
\email{petya}{bultreebank.org}
\and
Kiril Simov$^{\S}$\\
\email{kivs}{bultreebank.org}
}

\date{} 
\maketitle 



\thispagestyle{empty}	     \pagestyle{empty}

\begin{abstract}
The paper presents a feature-rich approach to the automatic recognition and categorization of named entities (persons, organizations, locations, and miscellaneous) in news text for Bulgarian. We combine well-established features used for other languages with language-specific lexical, syntactic and morphological information. In particular, we make use of the rich tagset annotation of the BulTreeBank (680 morpho-syntactic tags), from which we derive suitable task-specific tagsets (local and nonlocal). We further add domain-specific gazetteers and additional unlabeled data, achieving F$_1$=89.4\%, which is comparable to the state-of-the-art results for English.
\end{abstract}



\section{Introduction}
\label{intro}
The earliest work on named entity recognition (\NER) was based on hand-crafted rules
using pattern matching~\cite{appelt:muc6}.
For instance, a rule could encode the knowledge that
a sequence of capitalized words ending in `\emph{Inc.}'
is typically the name of an organization.
An example of such a system is \ANNIE\ in the \GATE\ architecture~\cite{Cun02b}.
Such systems could achieve very high precision, but typically suffered from low
recall.  They also required significant manual tuning, which was time-consuming
and could be quite complicated when thousands of rules interact in complex
manners.

Since the nineties, statistical models have offered a viable alternative
while requiring little or no manual tuning at all.
Such models typically treat \NER\ as a sequence tagging problem,
where each word is tagged with its entity type if it is part of an entity.
Generative models such as Hidden Markov Models (\HMM)~\cite{bikel09, zhou01}
have demonstrated excellent performance on the
\emph{Message Understanding Conference} (\MUC)
datasets~\cite{chinchor98}.

Discriminative models such as locally-normalized maximum-entropy~\cite{Borthwick99}
and conditional random fields (\CRF)~\cite{McCallum:03b} have also been
explored for \NER.
Collins~\cite{Collins02} used an \HMM\ tagger
to generate $n$-best outputs, which he reranked discriminatively.
By using a semi-Markov \CRF,
\cite{Sarawagi04}
recast \NER\ as a segmentation rather than a tagging problem,
thus allowing for richer feature sets.

Recent research also includes semi-supervised methods,
e.g., \cite{Miller04} use word clusters
derived from large sets of unlabeled data
in order to enrich their feature set.

\NER\ can also be viewed as a two-stage process:
(1) find the named entities in a sentence,
and (2) classify each entity into its type, e.g, person, organization,
location, etc. \cite{Collins02} mentions that first identifying named
entities without classifying them alleviates some data sparsity issues.
\cite{Collins99} focus on the second stage,
named entity classification,
assuming that the boundaries of the named entities have been found already;
they use a bootstrapping approach
based on co-training in order to leverage unlabeled examples.
\cite{Riloff99} use a similar bootstrapping approach
for information extraction.

Using \CRF s has become the dominant approach to \NER~\cite{McCallum:03b},
allowing for effective feature construction,
handling very large feature sets,
and modeling complex interactions across multiple levels of granularity;
Thus, in the present paper, \CRF s
will be our learning method of choice.
We will employ a rich set of features that
(\emph{i}) have been found useful for other languages,
(\emph{ii}) can handle expert knowledge in the form of gazetteers and domain-specific predicates,
(\emph{iii}) can model rich morpho-syntactic characteristics of Bulgarian,
(\emph{iv}) can represent complex predicates,
(\emph{v}) can be extracted from raw text automatically.

In the remainder of the paper, we will describe feature generation,
and we will discuss the results.


\section{Sequence tagging model}

The identification of named entity mentions in text can be implemented using a
sequence tagger, where each token is labeled with a BIO tag
indicating whether it begins (B), is inside (I), or is outside (O)
of a named entity mention~\cite{bio95}.
Following CoNLL-2002~\cite{tksintro2002conll},
we further indicate whether it is a
person (PER), an organization (ORG), a location (LOC), or miscellaneous (MISC),
which yields the following nine tags:
B-PER, I-PER, B-ORG, I-ORG, B-LOC, I-LOC, B-MISC, I-MISC, and O.
See Figure \ref{fig:example} for an example.

\begin{figure}
\centering
{\small \begin{tabular}{@{}c@{ }@{ }c@{ }@{ }c@{ }@{ }c@{ }@{ }c@{ }@{ }c@{ }@{ }c@{}}
\textcyrrm{Georgi} & \textcyrrm{P\cdprime rvanov} & \textcyrrm{e} & \textcyrrm{prezident} & \textcyrrm{na} & \textcyrrm{B\cdprime lgariya} & .\\
\emph{Georgi} & \emph{Parvanov} & \emph{is} & \emph{President} & \emph{of} & \emph{Bulgaria} & . \\
B-PER  & I-PER    &  O & O       &  O &  B-ORG & O\\
\end{tabular}}
\caption{Sample tagging using the BIO tagset.
The sentence contains two named entity mentions:
one person, and one organization.}
\label{fig:example}
\end{figure}

\section{Feature sets}

\subsection{Basic features}

Feature-based models like \CRF s are attractive since they reduce
the problem to finding a feature set that adequately represents
the target task. We used features based on individual words
as well as orthographic predicates, as shown in Table \ref{tab:orth},
and character-level $n$-gram predicates, $2 \leq n \leq 4$.
Bulgarian is morphologically rich and thus such predicates can help the system
recognize informative substrings
in words that were not seen in training, e.g.,
Slavic family name endings like \textcyrit{-ov} (`\emph{-ov}')
or \textcyrit{-ova} (`\emph{-ova}'),
and the possessive \textcyrit{-vo} (`\emph{-vo}') that is often used at the end of
Bulgarian location names.
We also included word prefix and
word suffix predicates, also of lengths $2 \leq n \leq 4$. This may seem
redundant, but prefix and suffix predicates also take into account the
position of the $n$-gram in the word, which can often be informative.
For example \textcyrit{vo} (`\emph{vo}') occurring at the end of a word is much more
informative than its presence anywhere in the word,
e.g., compare \textcyrit{\underline{vo}da} (`\emph{voda}', i.e., `water')
vs. \textcyrit{Dimitro\underline{vo}} (`\emph{Dimitrovo}', a village name).
We further included predicates that
indicate whether the current token occurs in parentheses
or inside a quotation.
Finally, in addition to the current token,
we also used features on the previous and on the following one.

Table \ref{tab:orth} lists the orthographic predicates we used.

\begin{table}[htb]
\centering
{\small \begin{tabular}{rll}
\bf Predicate name & \bf Regular expression\\
\hline
Initial capital & [\textcyrrm{A-YA}].*\\
Capital, followed by any & [\textcyrrm{A-YA}]. \\
Initial capitals, alpha & [\textcyrrm{A-YA}][\textcyrrm{a-ya}]*\\
All capitals & [\textcyrrm{A-YA}]+ \\
All lowercase & [\textcyrrm{a-ya}]+ \\
Capitals mix & [\textcyrrm{A-YAa-ya}]+ \\
Contains a digit & .*[0-9].* \\
Single digit & [0-9] \\
Double digit & [0-9][0-9] \\
Natural number & [0-9]+\\
Real number & [-0-9]+[\verb!\!.,]?[0-9]+\\
Alpha-numeric & [\textcyrrm{A-YAa-ya}0-9]+\\
Roman & [ivxdlcm]+$ | $[IVXDLCM]+ \\
Contains dash & .*-.* \\
Initial dash & -.* \\
Ends with dash & .*- \\
Punctuation & [,\verb!\!.;:\verb!\!?!-+\verb!"!] \\
Multidots & \verb!\!.\verb!\!.+ \\
Ends with dot & .*\verb!\!.\\
Acronym & [\textcyrrm{A-YA}]+ \\
Lonely initial & [\textcyrrm{A-YA}]\verb!\!. \\
Single character & [\textcyrrm{A-YAa-ya}]\\
Quote & [\verb!"!\verb!'!]
\end{tabular}}
\caption{
\label{tab:orth}
\textbf{The orthographic predicates used in our system.}
The observation list for each token will include a
predicate for each regular expression that matches it.
}
\end{table}

Even though simple, the above feature set
yielded a very good performance on the development data:
see Table~\ref{tab:results}, row A.
In order to add expert knowledge to
the model, we used some regular expressions
that generate predicates on the word by checking whether
it ends with some character sequences that are common for Bulgarian names of persons,
e.g., \textcyrit{-ska} (`\emph{-ska}'), \textcyrit{-ski} (`\emph{-ski}'),
\textcyrit{-ov} (`\emph{-ov}'), \textcyrit{-va} (`\emph{-va}')
or locations, e.g., \textcyrit{-vo} (`\emph{-vo}');
see Table~\ref{tab:results}, row B for details.
Another useful approach for adding
domain knowledge to the model, used in previous work for named entity recognition
and gene mentions tagging, is predicate generation on the basis of
membership in a gazetteer.

In our case, gazetteers are lists of words,
multi-token units, and acronyms.
A straightforward method of integrating these knowledge sources
is to create predicates indicating whether a token
occurs in one of these gazetteers. For multi-token entries, we required
that all entry tokens be matched.
We further created similar predicates for the previous and the next tokens. Table~\ref{tab:results}, row C summarizes
the effect of adding these gazetteers.

The following sections will further explore the morpho-syntactic tagset,
the feature induction and using additional raw unannotated text.

\begin{table*}[htb]
\centering
{\small \begin{tabular}{lllrrr}
& \bf Types of predicates & \bf NE type & \bf Precision & \bf Recall & \bf F$_1$-Measure\\
\hline
{\bf A} & Orthographic & Organization & 85.50 & 82.73 & 84.10 \\
& & Person & 86.05 & 79.86 & 82.84 \\
& & Location & 88.34 &  82.97 &  85.57 \\
& & Miscellaneous & 44.83 & 22.41 & 29.89 \\
& & \bf OVERALL & 85.67 & 78.89 & \bf 82.14\\
\hline
{\bf B} & +Domain-specific & Organization & 85.35 & 83.81 & 84.57 \\
& & Person & 86.46 & 80.40 & 83.32 \\
& & Location & 88.51 & 82.48 & 85.39 \\
& & Miscellaneous & 44.83 & 22.41 & 29.89 \\
& & \bf OVERALL & 85.86 & 79.20 & \bf 82.40 \\
\hline
{\bf C} & +Gazetteers & Organization & 87.89 & 80.94 & 84.27 \\
& & Person & 90.70 & 84.17 & 87.31 \\
& & Location &  88.45 & 87.59 & 88.02 \\
& & Miscellaneous & 48.39 & 25.86 & 33.71 \\
& & \bf OVERALL & 88.26 & 81.96 &  \bf 85.00 \\
\hline
{\bf D} & +Local morphology & Organization & 88.93 & 86.69 & 87.80 \\
& & Person & 92.96 & 90.13 & 91.52 \\
& & Location &  89.64 & 90.29 & 89.96 \\
& & Miscellaneous & 57.14 & 27.12 & 36.78 \\
& & \bf OVERALL & 90.19 & 86.60 & \bf 88.36 \\
\hline
{\bf E} & +Nonlocal morphology & Organization & 87.23 & 88.49 & 87.86 \\
& & Person & 90.99 & 92.46 & 91.72 \\
& & Location & 90.34 & 90.78 & 90.56 \\
& & Miscellaneous & 60.00 & 25.42 & 35.71 \\
& & \bf OVERALL & 89.36 & 88.06 & \bf 88.70 \\
\hline
{\bf F} & +Feature induction & Organization & 89.45 & 88.49 & 88.97 \\
& & Person & 93.13 & 92.46 & 92.79 \\
& & Location & 88.11 & 91.75 & 89.89 \\
& & Miscellaneous & 60.00 & 25.42 & 35.71 \\
& & \bf OVERALL & 90.02 & 88.36 & \bf 89.18 \\
\hline
{\bf G} & +Mutual information & Organization & 89.89 & 89.57 & 89.73 \\
& & Person & 93.13 & 92.46 & 92.79 \\
& & Location & 88.89 & 91.26 & 90.06 \\
& & Miscellaneous & 60.00 & 25.42 & 35.71 \\
& & \bf OVERALL & 90.38 & 88.44 & \bf 89.40
\end{tabular}}
\caption{\textbf{Precision, recall and F$_1$-measure (in \%s) for different feature sets on the test dataset.}
(A) Uses orthographic predicates and some simple features like token length. We define this system as our baseline.
(B) Adds some simple regular expressions that match common patterns in Bulgarian personal and location names.
(C) Adds predicates for gazetteer membership.
(D) Adds predicates using local morpho-syntactic characteristics of the current word.
(E) Adds nonlocal morpho-syntactic characteristics.
(F) Adds feature induction to generate suitable combinations two of or more simple predicates.
(G) Further uses unlabeled text.}
\label{tab:results}
\end{table*}


\subsection{Morpho-syntactic features}



We made use of the rich tagset annotation
of the BulTreeBank~\cite{bul-morph},
from which we derived suitable task-specific tagsets (local and nonlocal).

Initially, we started with the full morpho-syntactic set of the BulTreeBank
(680 morpho-syntactic tags), and we were able to achieve some improvements.
However, working with so many distinct tags caused data sparsity issues,
and missed opportunities for successful generation.
We found the tagset was tightly coupled,
thus reducing the possibility to model complex context relationships in the text sequence.
Some of the tags were quite rare and apparently not very helpful.
Since our \NER\ experiments aim to be practical,
we divided the tag characteristics (morpho-syntactic and part of speech)
into \emph{local} and \emph{nonlocal}.
%
%
The \emph{local} predicates (111 tags in this set) are linguistically
related to other predicates that hold on the same word,
e.g., character $n$-grams, prefixes and suffixes,
the word itself, etc.
For nouns, they could be gender (e.g., masculine, feminine, neuter),
number (e.g., singular, plural, count form),
article (e.g., indefinite, definite).
The \emph{nonlocal} predicates (230 tags in this set)
are related to predicates that hold on words in a particular context,
i.e., window around the target word, e.g.,
the type of noun: common vs. proper.


This treatment of the BulTreeBank tagset
stimulates simple and adequate treatment of the feature functions
design for the \NER\ task.
The \emph{local} characteristics are used alone and in
combination with predicates holding on the current word,
while the \emph{nonlocal} ones are combined with predicates and words
appearing in the local context at positions $\{-3, -2, -1, 0, +1, +2, +3\}$,
where $0$ is the current word, $-1$ is the previous one, $+1$ is the next one, etc.
This approach induces many useful predicates,
which is shown by the overall increase in the system performance:
see Table \ref{tab:results}, rows D and E.

For instance, the following feature could be useful:

\begin{equation}
f_i(s,o)=\begin{cases}
\mathbf{1} & \mathbf{if} \; '\mathrm{WORD}=\textcyrit{Dzhina}' \; \in o,\\
& '\mathrm{local\_tag}=\mathrm{N-msi}' \; \in o,\\
& \mathbf{tag_0}(s)=\mathrm{B-PER};\\
\mathbf{0} & \mathbf{otherwise}.
\end{cases}
\end{equation}

In the above example, the feature function will have the value of 1
if the the word is \textcyrit{Dzhina} (`\emph{Dzhina}') and its local tag characteristics are
feminine, singular, indefinite, and the named entity tag at this position is 'B-PER';
otherwise, the function value will be 0.

In contrast, we show that \emph{nonlocal} tags would be beneficial
in modeling complex context dependencies, for example:

\begin{equation*}
f_i(s,o)=\begin{cases}
\mathbf{1} & \mathbf{if} \; '\mathrm{WORD_{+2}}=\textcyrit{vleze}' \; \in o,\\
& \mathrm{'nonlocal\_tag=p'} \; \in o,\\
& \mathbf{tag_0}(s)=\mathrm{B-PER};\\
\mathbf{0} & \mathbf{otherwise}.
\end{cases}
\end{equation*}

In the above example,
the function value will be $1$ if the nonlocal tag describes a proper noun,
the word \textcyrit{vleze} (`\emph{entered}')
appears at position $+2$, and the current tag is `B-PER'.
In all other cases, it will be $0$.

The BulTreeBank tagset was further reduced by dropping information
about the types of pronouns,
the article was limited to indefinite and definite only,
and the number and the count forms were merged into a single class.
Rows D and E in Table \ref{tab:results} show the results
when using these morpho-syntactic features.



\subsection{Inducing complex features}

So far, we have described features over a single predicate only, except
in the design of morpho-syntactic features.
However, it is often useful to create features based on the conjunction
of several simple predicates:

\begin{equation*}
f_i(s,o)=\begin{cases}
\mathbf{1} & \mathbf{if} \; '\mathrm{WORD}=\textcyrit{Batenberg}' \; \in o,\\
& '\mathrm{WORD_{+1}}=\textcyrit{upravlyava}' \; \in o,\\
& \mathbf{tag_0}(s)=\mathrm{B-PER};\\
\mathbf{0} & \mathbf{otherwise}.
\end{cases}
\end{equation*}

The above feature could be useful for disambiguating the token
\textcyrit{Batenberg}\footnote{Alexander Joseph of Battenberg
(April 5, 1857 -- October 23, 1893) was the first prince (knyaz) of modern Bulgaria.} (`\emph{Batenberg}'),
which can be a person's name (e.g., when followed by \textcyrit{upravlyava}, `\emph{ruled}').
However, it could be also part of a location (e.g., when preceded by \textcyrit{ploshchad},
`\emph{a square}'),
and thus we might want to have a special feature for this case:
\begin{equation*}
f_i(s,o)=\begin{cases}
\mathbf{1} & \mathbf{if} \; '\mathrm{WORD}=\textcyrit{Batenberg}' \; \in o,\\
& '\mathrm{WORD_{-1}}=\textcyrit{ploshchad}' \; \in o,\\
& \mathbf{tag_0}(s)=\mathrm{B-LOC};\\
\mathbf{0} & \mathbf{otherwise}.
\end{cases}
\end{equation*}

The system already uses tens of thousands of features,
which makes it infeasible to create predicates
for the conjunction of all pairs of simple predicates.
Even if it were computationally possible,
it would still be hard to gather sufficient statistics for most of them.
Thus, we use the method described in \cite{McCallum:03}
to limit the search space.
Row F in Table \ref{tab:results} shows the results.

\subsection{Using unlabeled text}

In this section, we try using additional unlabeled text,
from which we extract two kinds of additional features.

The first type is pointwise \emph{mutual information} (\MI).
It is a standard measure of the strength of association between
co-occurring items and has been used successfully in extracting collocations from
English text~\cite{Lin98} and for performing Chinese word segmentation~\cite{Riloff99, Maosong98, Zhang02}, among other tasks.

The \MI\ for two words $x_1$ and $x_2$ is defined as follows:

\begin{equation*}
 \mathrm{MI}(x_1,x_2) = \log \frac{\mathrm{Pr}(x_1,x_2)}{\mathrm{Pr}(x_1)\mathrm{Pr}(x_2)}
\end{equation*}

\noindent where $\mathrm{Pr}(x)$ is the probability of observing $x$,
and $\mathrm{Pr}(x_1,x_2)$ is the probability of $x_2$ following $x_1$.

Estimates of the \MI\ are simple and cheap to compute from unlabeled data alone;
this can be done in linear time on the text length.

We used 7.4M words of unlabeled newswire text,
from which we extracted the top
1M word pairs, ranked according to the \MI\ score.
We then distributed these pairs into separate bins based on their \MI\ values,
where bins contained approximately equal numbers of pairs,
and we created binary features of the following kind to be integrated in the \CRF\ model:


\begin{equation*}
f_i(s,o)=\begin{cases}
\mathbf{1} & \mathbf{if} \; '\mathrm{WORD}=\textcyrit{Batenberg}' \; \in o,\\
& '\mathrm{WORD_{-1}}=\textcyrit{ploshchad}' \; \in o,\\
& \mathrm{MI(WORD,WORD_{+1}) \in bin_x},\\
& \mathbf{tag_0}(s)=\mathrm{B-LOC};\\
\mathbf{0} & \mathbf{otherwise}.
\end{cases}
\end{equation*}

Initially, we tried using a high number of bins (50K, 100K, 200K and 500K),
but did not observe improvements on the development set,
probably because of the limited amount of unlabeled text
and the sparsity issues resulting thereof.
We thus tried smaller numbers of bins, eventually ending up with just two bins,
which yielded the highest improvement on the development set.
Table 2, row G shows this also yielded a tiny improvement on the test set:
from 89.18\% to 89.40\%.

We also tried a second kind of features based
on the clustering algorithm described in \cite{Brown92},
using (1) bottom-up agglomerative word clustering,
and (2) the clustering method of \cite{Liang:thesis},
but were unable to achieve any performance gains on the development dataset.


\section{Experiments and evaluation}



In our experiments, we used the Mallet implementation of \CRF.
We further used manually annotated sentences from the BulTreeBank
for training, development and testing:
\begin{itemize}
  \item \emph{training}: 8,896 sentences (147,339 tokens), including
    1,563 organizations, 4,282 persons, 2,300 locations, and 353 miscellaneous named entities;
  \item \emph{development}: 1,779 sentences (29,467 tokens), including
    312 organizations, 856 persons, 383 locations, and 70 miscellaneous named entities;
  \item \emph{testing}: 2,000 sentences (34,649 tokens), including
    315 organizations, 841 persons, 438 locations, and 69 miscellaneous named entities.
\end{itemize}

In the process of system development,
we did many iterations of training and evaluation on the development data,
followed by predicate enhancement and new feature construction.
For the final evaluation,
we trained on a concatenation of the training and the development data
and we tested on the unseen test data.

We were very strict in the evaluation and gave no credit
for partially discovered named entities:
we considered that a named entity
was correctly recognized if all tokens it covers were labeled correctly,
and no extra tokens were included as part of the entity.

\section{Results and discussion}


The evaluation results are shown in Table \ref{tab:results}.
We started with simple orthographic features in our baseline system (row A: F$_1$=82.14\%),
and we repeatedly added additional types to improve the performance.

As we can see in rows B and C,
using domain-specific features
in the form of simple regular expressions and gazetteers
yielded about 3\% absolute improvement on F$_1$ to 85\%.
Adding morpho-syntactic features resulted in additional 3\% increase to 88.70\%
(rows D and E), and using feature induction and mutual information (rows F and G)
added 1\% more to F$_1$, which reached 89.40\% for our final system (row G).

An examination of system's output on the development dataset
shows that the primary source of errors were properly labeled mentions
whose boundaries were off by one or more tokens.
If the score was relaxed so that tagged entities were considered as true positives
if and only if one or more tokens overlap with a correct entry,
the performance on the development data would increase a lot.
As an extreme example, consider the string \textcyrit{Vashku da Gama}
(`\emph{Vashku da Gama}', i.e., `\emph{Vasco da Gama}'),
which was incorrectly recognized as covering two entities of type person
(`\textcyrit{Vashku}' and `\textcyrit{Gama}').


We should note that our results are not directly comparable to previous publications;
we are the first to try a statistical approach for Bulgarian \NER,
which has attracted very little research interest so far
and was dominated by rule-based systems.
For example, \cite{bul-ner} describe adding manual rules for Bulgarian \NER\ to \ANNIE,
but provide no formal evaluation.

It is still informative to compare our results
to those achieved for other Slavic languages
even if we use different kinds/amounts of training data
and different sets of named entities types.
For example, the best P/R/F$_1$ results for Russian are 
79.9/63.7/70.9 (in \%),
which was calculated for six types of named entities \cite{Popov:al:NER:Russian}:
persons (70.5/53.9/61.1),
organizations (72.5/59.8/65.5),
locations (91.2/68.7/78.4),
dates (77.0/71.7/74.3),
percents (87.5/87.5/87.5),
and money (80.8/40.4/60.6).
For Polish, the best results are the following \cite{Piskorski:2004:NER:Polish}:
persons (90.6/85.3/87.9),
organizations (87.9/56.6/68.9),
locations (88.4/43.4/58.2),
time (81.3/85.9/83.5),
percents (100.0/100.0/100.0),
and money (97.8/93.8/95.8);
overall this makes 91.0/77.5/82.4.
For Czech, the best results are the following \cite{kravalova-zabokrtsky:2009:NEWS}:
84.0/70.0/76.0.
We can see that our results are superior, especially on F$_1$ and recall.

The state-of-the-art F$_1$ scores for English at specialized competitions
like the Message Understanding Conference and CoNLL-2003
has been 93.39\% and 88.76\%, respectively.
Similarly, at CoNLL-2002\footnote{\texttt{http://www.cnts.ua.ac.be/conll2002/ner/}} and CoNLL-2003\footnote{\texttt{http://www.cnts.ua.ac.be/conll2003/ner/}},
the best F$_1$ for German, Spanish and Dutch
were 72.41\%, 81.39\% and 77.05\%, respectively.
The highest reported F$_1$ score for Arabic,
which is morphologically richer than Bulgarian,
is 83.5\%~\cite{arab:ner}.
All these systems were trained on about 200K tokens as is ours,
and thus we can conclude that our F$_1$=89.4\% is comparable
to the state-of-the-art.




\section{Conclusions and future work}


Our experiments show that \CRF\ models with carefully-designed features
can identify mentions of named entities
(organizations, persons, locations and miscellaneous)
in Bulgarian text with fairly high accuracy,
even without features containing domain-specific knowledge.
However, such features,
which in our framework take the form of membership in a gazetteer,
simple common endings for personal names and location entities,
and rich morpho-syntactic tagsets, can lead to improved system performance.
Even on the limited training data we had available,
we have shown that using external raw text could potentially help on the system performance.
However, broader experiments are needed to measure the scope of influence.

We also demonstrate that proper handling of morpho-syntactic tags
for morphologically rich languages like Bulgarian could lead to intelligent
feature generation and huge performance gains for the named entity tagger.
Still, we consider the construction of morpho-syntactic taggers
that can handle the rich tagset of the BulTreeBank as a challenging but demanding task.
Finally, using raw text is another promising direction we plan to pursue in future work.





\bibliographystyle{abbrv}  
\begin{scriptsize}
\bibliography{ranlp-biblio}  
\end{scriptsize}

\end{document}